\documentclass{article}
\usepackage{ijcai25}

\usepackage{times}
\usepackage{soul}
\usepackage{url}
\usepackage[hidelinks]{hyperref}
\usepackage[utf8]{inputenc}
\usepackage[small]{caption}
\usepackage{graphicx}
\usepackage{amsmath}
\usepackage{amsthm}
\usepackage{booktabs}
\usepackage{algorithm}
\usepackage{algorithmic}
\usepackage[switch]{lineno}

\usepackage{array}
\usepackage{multirow}
\usepackage{subcaption}
\usepackage{makecell}
\usepackage{xcolor}

\title{GRATR: Zero-Shot Evidence Graph Retrieval-Augmented Trustworthiness Reasoning}

\author{
Ying Zhu$^1$
\and
Shengchang Li$^2$
\and
Ziqian Kong$^{3}$\and
Qiang Yang$^1$\And
Peilan Xu$^1$ \thanks{Corresponding author.}\\
\affiliations
$^1$Nanjing University of Information Science and Technology\\
$^2$The University of Melbourne\\
$^3$Hangzhou Dianzi University\\
\emails
202283460055@nuist.edu.cn,
is.lishch@gmail.com,
kzq@hdu.edu.cn,
mmmyq@126.com,
xpl@nuist.edu.cn
}

\begin{document}

\maketitle
 
\begin{abstract}

Trustworthiness reasoning aims to enable agents in multiplayer games with incomplete information to identify potential allies and adversaries, thereby enhancing decision-making. In this paper, we introduce the graph retrieval-augmented trustworthiness reasoning (GRATR) framework, which retrieves observable evidence from the game environment to inform decision-making by large language models (LLMs) without requiring additional training, making it a zero-shot approach. Within the GRATR framework, agents first observe the actions of other players and evaluate the resulting shifts in inter-player trust, constructing a corresponding trustworthiness graph. During decision-making, the agent performs multi-hop retrieval to evaluate trustworthiness toward a specific target, where evidence chains are retrieved from multiple trusted sources to form a comprehensive assessment. Experiments in the multiplayer game \emph{Werewolf} demonstrate that GRATR outperforms the alternatives, improving reasoning accuracy by 50.5\% and reducing hallucination by 30.6\% compared to the baseline method. Additionally, when tested on a dataset of Twitter tweets during the U.S. election period, GRATR surpasses the baseline method by 10.4\% in accuracy, highlighting its potential in real-world applications such as intent analysis.

\end{abstract}

\section{Introduction}

\begin{figure}[htbp]
\centering
\includegraphics[width=0.45\textwidth]{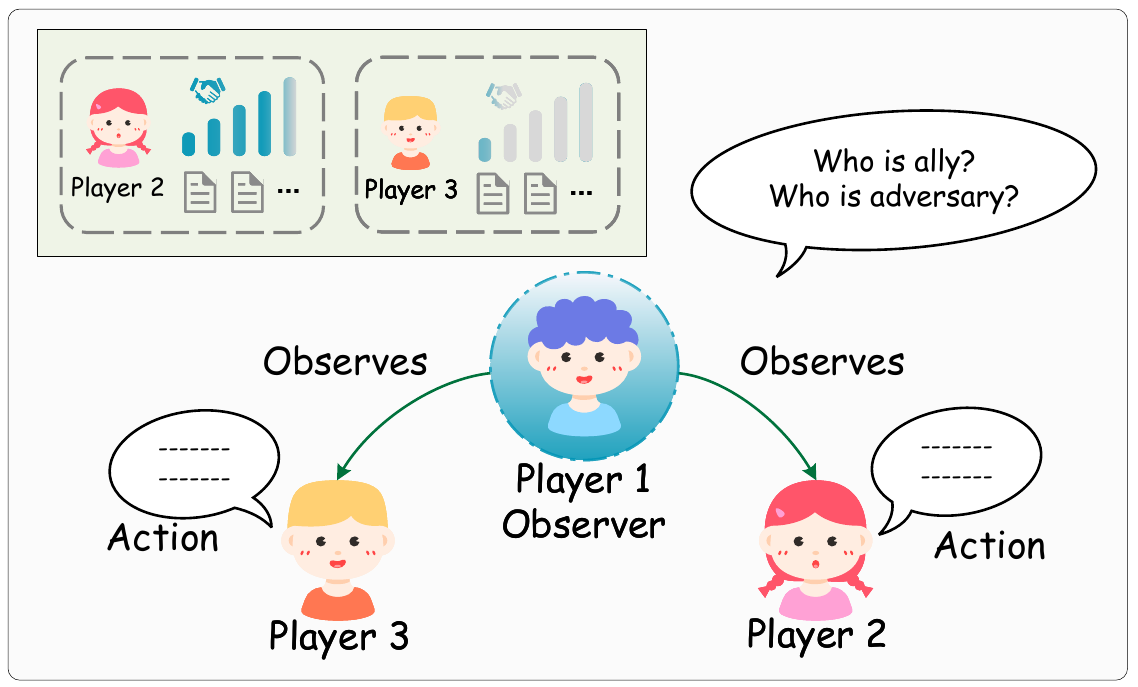}
\caption{Illustration of trustworthiness reasoning. Agent observes the actions of other players to gather evidence, and then evaluates inter-player trust and informs decision-making.}
\label{fig:tr}
\end{figure}

In multiplayer games with incomplete information, trustworthiness reasoning is critical for evaluating the intentions of players, who may conceal their true motives through actions, dialogue, and other observable behaviors. Autonomous agents analyze the trustworthiness of players based on observable actions to identify potential allies and adversaries (Fig. \ref{fig:tr}). Current methods supporting such reasoning include symbolic reasoning \cite{nye2021improving,lu2021inter}, evidential theory \cite{liu2021double}, Bayesian reasoning \cite{wojtowicz2020probability,sohn2021neural,wan2021gaussianpath}, and reinforcement learning \cite{wan2021reasoning,wang2020adrl,tiwari2021dapath}. While effective, these methods struggle to address the complexity of natural language interactions, the ambiguity of player behavior, and the dynamic nature of strategic decision-making in such environments.

To address these limitations, large language models (LLMs) offer a promising approach for trustworthiness reasoning in multiplayer games, owing to their advanced natural language understanding and generation capabilities \cite{brown2020language,kenton2019bert,radford2019language}. LLMs utilize these capabilities to interpret complex dialogues, infer latent intentions, and detect deceptive behaviors from contextual cues. However, LLMs face inherent challenges, including the risk of hallucination and knowledge obsolescence \cite{ji2023survey,maynez2020faithfulness}. To mitigate these issues, techniques such as supervised fine-tuning and reinforcement learning have been proposed to enhance their reasoning performance \cite{ouyang2022training,stiennon2020learning}. Nonetheless, these approaches often require extensive historical data and well-defined reward signals, which may be scarce or unavailable in real-world game scenarios.

To enhance the capabilities of LLMs in dynamic, knowledge-intensive environments, retrieval-augmented generation (RAG) \cite{gao_retrieval-augmented_2024,zhao_retrieval-augmented_2024} has emerged as a promising alternative. RAG addresses the limitations of LLMs by integrating an external retrieval mechanism that dynamically fetches relevant information to augment the generation process. In the RAG framework, a retriever first indexes and retrieves pertinent data chunks, which are then combined with an input query to refine the generation process. This approach mitigates issues such as knowledge obsolescence and hallucination by incorporating up-to-date and contextually relevant information, making it a promising solution for trustworthiness reasoning in multiplayer games with incomplete information.

However, trustworthiness reasoning in multiplayer games presents additional challenges that exceed the capabilities of current RAG methods. Specifically, it requires the real-time collection and analysis of statements and actions as evidence exhibited by players. Due to the complexity of player interactions, trustworthiness reasoning for a given player must consider the actions of other players toward that target. This necessitates multi-hop retrieval and synthesis of evidence, which becomes computationally intensive and time-consuming, particularly in scenarios involving many players.

\textbf{Contributions.} We propose a novel method, the graph retrieval-augmented trustworthiness reasoning (GRATR) framework, which constructs a dynamic trustworthiness graph to model player interactions in real time, thus avoiding the computational overhead of retrieving information from large text corpora repeatedly. During the observation phase, agents collect observable evidence to dynamically update the graph's nodes (representing players) and edges (representing trust relationships). During decision-making, GRATR performs multi-hop retrieval to evaluate the trustworthiness of a specific target player, leveraging evidence chains from multiple trusted sources to form a comprehensive assessment. This approach enhances reasoning and decision-making without additional training, making it a zero-shot solution. We validate GRATR in the multiplayer game \emph{Werewolf}, comparing it to baseline LLMs and LLMs with state-of-the-art RAG techniques. The experimental results demonstrate its ability to model dynamic trust relationships and support informed decision-making in complex, incomplete information scenarios. Furthermore, GRATR enhances transparency and traceability by visualizing temporal evidence and evidence chains through the trustworthiness graph, overcoming the limitations of previous methods. Beyond multiplayer games, we also apply GRATR to real-world scenarios, i.e., analyzing the intent behind social media tweets, showcasing its broader applicability.

\section{Preliminary}
In a multi-player game with incomplete information, the game can be described by the following components:

\begin{itemize}
	
	\item \textbf{Players}: $P = \{p_1, p_2, \ldots, p_n\}$, where $p_i$ represents the $i$-th player, and each player $p_i$ has a private type $\theta_i \in \Theta_i$, where $\Theta_i$ is the set of possible types for player $p_i$.
	
	\item \textbf{Actions}: In each round $t$, player $p_i$ chooses an action $a_i^t \in A_i$, where $A_i$ is the set of available actions for player $p_i$.
	
	\item \textbf{Observations}: After all players choose their actions, each player $p_i$ receives an observation $ o_i^t \in O_i$, where $O_i$ is the set of possible observations for $p_i$. The observation $o_i^t$ depends on the joint actions $\mathbf{a}^t = (a_1^t, a_2^t, \ldots, a_n^t)$ and possibly other public or private signals.
	
	\item \textbf{Objective}: Each player $p_i$ aims to maximize their expected utility $ \mathbf{E}_{\sigma_i}[u_i(\mathbf{a}, \theta)]$, where the expectation is taken over the belief distribution $ \sigma_i$.
	
	\item \textbf{Game Dynamics}: The game proceeds as follows:
	\begin{itemize}
		\item At the beginning of each round $t$, each player $ p_i$ observes $h_i^t$ and selects an action $a_i^t = s_i(h_i^t, \theta_i)$.
		\item After all actions $\mathbf{a}^t$ are chosen, players receive observations $o_i^t$.
		\item Players update their beliefs $ \sigma_i^{t+1}(\theta_{-i} \mid h_i^{t+1})$ based on the new history $h_i^{t+1}$ that includes $o_i^t$ and $a_i^t$.
		\item The game continues for a fixed number of rounds $T$, or until a stopping condition is met.
	\end{itemize}
	
\end{itemize}

\section{Methodology}

\begin{figure*}[h]
	\centering
	\includegraphics[width=\textwidth]{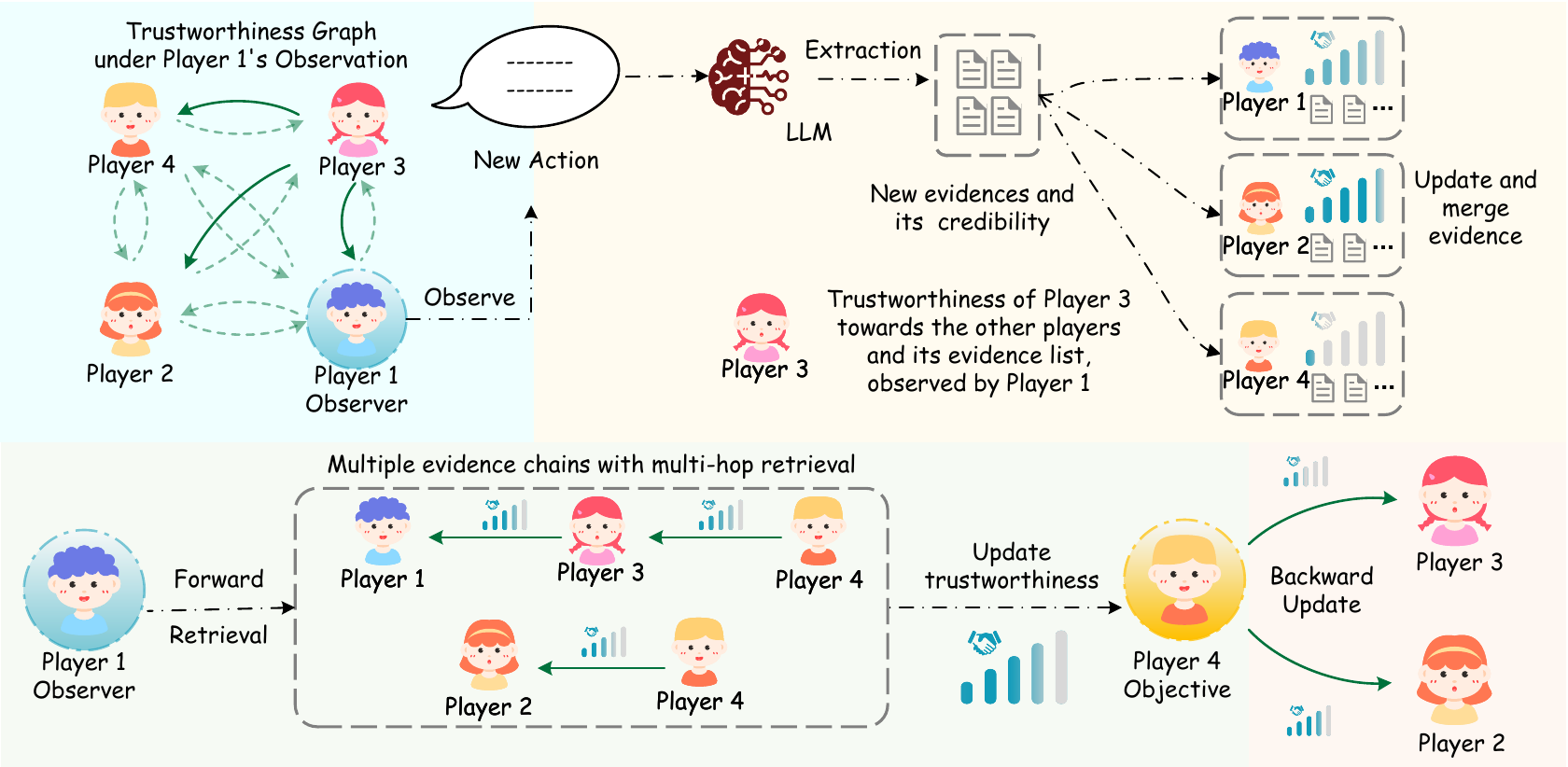}
	\caption{The overall framework of GRATR: Step 1. An agent participates in the game as player 1, and initializes a trustworthiness graph $G$. Step 2. When player 1 receives a new observation $o^t_1(p_3)$ following an action by $a^t_3$ at time $t$, it uses an LLM to extract the action into new evidence and its credibility and then updates and merges evidence on the graph $G^t$. Step 3. Player 1 obtains multiple evidence chains by multi-hop retrieval and updates the trustworthiness of player 4. Step 4. Update the trustworthiness of player 4 towards player 2 and player 3.}
	\label{fig:fr}
\end{figure*}

To enhance the effectiveness of LLM reasoning, especially in environments where trust and strategic interactions are crucial, it is essential to retrieve the most relevant evidence from historical data. This motivated us to develop a framework where the information observed by agents is structured into a graph-based evidence base. By maintaining this evidence graph, we can retrieve related evidence chains, augment LLM reasoning, and mitigate the issues of hallucination and opacity. This methodology forms the foundation of our proposed GRATR system. Figure \ref{fig:fr} presents the framework of GRATR. The process begins with the initialization of a trustworthiness graph when an agent participates in the game. Observations made by the agent are analyzed using the LLM to extract evidence and assess its credibility, which is then used to update the trustworthiness graph $G$. Through multi-hop retrieval on $G$, evidence chains are constructed to evaluate the trustworthiness of other players. Finally, the system updates trustworthiness relationships among players based on the gathered evidence, leveraging the graph structure to provide a transparent and well-grounded reasoning process.

\subsection{Initialization of the Trustworthiness Graph}

Assume an agent participates as a player in a multiplayer game with incomplete information, maintaining a directed graph $G^t$ to record historical observations $h^t$ up to round $t$ as a dynamic evidence base. This graph $G^t$ serves as the foundation for the agent's reasoning process, enabling the retrieval and use of real-time evidence. The graph consists of two core components: nodes and edges.

\textbf{Nodes:} Each node in the graph $G^t$ represents a player $p_i$ and stores two parameters.

\begin{itemize}
    \item \textbf{Trustworthiness of Nodes} $T^t(p_i) \in [-1,1]$: The perceived trustworthiness of player $p_i$ by the agent at time $t$. When $T^t(p_i) > \epsilon$, the agent regards player $p_i$ as an ally, when $T^t(p_i) < -\epsilon$, the agent regards player $p_i$ an adversary; otherwise, the agent regards player $p_i$ as indifferent.

    \item \textbf{Historical Observations} $h^t(p_i)$: The history of observations gathered by the agent about player $p_i$ up to round $t$, serving as the evidence base.
\end{itemize}

\textbf{Edges:} Each directed edge $e^t(p_i, p_j)$ connects player node $p_i$ to player node $p_j$ and stores two parameters.

\begin{itemize}
    \item \textbf{Evidence List} $D^t(p_i, p_j)$: This list contains a set of evidence items $d^t(p_i, p_j)$ that record the actions of player $p_i$ towards player $p_j$ as observed by the agent. Each evidence item includes the specific action taken and its associated credibility $c^t(p_i, p_j)$, indicating the significance of this action in assessing trustworthiness.

    \item \textbf{Trustworthiness of Edges} $T^t(p_i, p_j)$: This weight reflects the agent's trust in player $p_i$ towards player $p_j$, determined by the accumulated evidence in the evidence list.
\end{itemize}

At the initial time $t=0$, the edge weight is set to zero, i.e., $T^0(p_i, p_j) = 0$, and the evidence list $D^0(p_i, p_j)$ is empty, indicating no prior observations or assessments.

\subsection{Update of the Trustworthiness Graph}

When the agent receives a new observation $o^t(p_i)$ following an action by player $p_i$, the evidence graph $G^t$ must be updated to incorporate this new information. This ensures that $G^t$ accurately represents the current state of trustworthiness among the players at time $t$.

The agent uses the LLM to extract evidence items $d^t(p_i, p_j)$ and their corresponding weights $c^t(p_i, p_j)$ from the observation $o^t(p_i)$ (the related prompt used for LLM interactions is provided in Appendix 1.1). For each directed edge $e^t(p_i, p_j)$ in the graph, the evidence list $D^{t+1}(p_i, p_j)$ associated with the edge $e^t(p_i, p_j)$ is updated by adding the new intention $d^t(p_i, p_j)$:

\begin{equation}\label{eqt:1}
	D^{t+1}(p_i, p_j) = D^t(p_i, p_j) \cup \{d^t(p_i, p_j)\}.
\end{equation}. 

The sign of $c^t(p_i, p_j)$ indicates the nature of $p_i$'s intention towards $p_j$: negative for hostility and positive for support, with $|c^t(p_i, p_j)|$ reflecting its strength. Note that the evidence list $D^t(p_i, p_j)$ is updated with the new observation, and the edge weight $T^t(p_i, p_j)$ is adjusted accordingly during retrieval to maintain an accurate representation of trustworthiness.

Meanwhile, the agent updates the trustworthiness of $p_j$ in response to the evidence items extracted from the LLM. The update depends on the two factors: the perceived trustworthiness of $p_i$, the credibility $c^t(p_i, p_j)$ of the  $d^t(p_i, p_j)$, which also represents the $p_i$'s confidence of the $p_j$'s current role classification $\mathcal{R}^t(p_j)$. The updated trustworthiness $T^t_i(p_j)$ is computed as follows:

\begin{equation}\label{eqt:2}
	u^t(p_j) = T^t(p_i) \cdot c^t(p_i, p_j),
\end{equation}, 

\begin{equation}\label{eqt:3}
T^{t+1}(p_j) =
\begin{cases}
T^{t}(p_j), & \text{if } |u^t(p_j)| \leq |T^{t}(p_j)|, \\
u^t(p_j), & \text{if } |u^t(p_j)| > |T^{t}(p_j)|,
\end{cases}
\end{equation}. 
where $u^t(p_j)$ represents the inference of $p_j$'s trustworthiness through the observation $o^t(p_j)$.

\begin{algorithm}[htbp]
	\caption{Graph Update Process}
	\label{alg:graph_update}
	\textbf{Input}: Graph $G^t$, observations $o^t$
	\begin{algorithmic}[1]
		\STATE Query the LLM to extract $\{\mathcal{R}^t(p_i, p_j)$, $d^t(p_i, p_j)$, $c^t(p_i, p_j)\}$ from the observation $o^t$;
		\FOR{each intention $d^t(p_i, p_j)$}
		\STATE Update the evidence list $D^{t+1}(p_i, p_j)$ using Eq. \eqref{eqt:1};
		\ENDFOR
		\FOR{each player $p_j$ connected by an edge $e^t(p_i, p_j)$}
		\STATE Update the trustworthiness $T^{t+1}(p_j)$ using the Eqs. \eqref{eqt:2}\eqref{eqt:3};
		\ENDFOR
	\end{algorithmic}
\end{algorithm}

\subsubsection{Evidence Merging}
In this phase, the objective is to aggregate and evaluate the various evidence collected by the agent over time, specifically related to the interactions between players $p_i$ and $p_j$. Assume that the agent has $n$ pieces of evidence $d^t(p_i, p_j)$ towards player $p_j$ in the evidence list $D^t(p_i, p_j)$ associated with the directed edge $e^t(p_i, p_j)$. The evidence is sorted in chronological order, with each piece of evidence having an associated weight $w^t(p_i, p_j)$ and a temporal importance factor $\rho$. The updated edge weight $\tau^{t+1}(p_i, p_j)$ is computed as follows:

\begin{equation}\label{eqt:4}
	T^{t+1}(p_i, p_j) = \tanh \left( \sum_{k=1}^{n} \rho^{n-k} \cdot c^t(p_i, p_j) \right)
\end{equation}, 

where the impact of evidence decreases over time, with more recent evidence having greater influence. The $\tanh$ function is used to constrain the edge weight $T^{t+1}(p_i, p_j)$ within the interval $[-1, 1]$, providing a bounded measure of the trustworthiness between players.

\subsection{Graph Retrieval Augmented Reasoning}

During the agent's turn, particularly when deciding on an action involving player $p_o$, the reasoning process is augmented by retrieving and leveraging relevant evidence from the evidence graph $G^t$. This graph-based retrieval augments the player's trustworthiness assessment by incorporating historical evidence into the reasoning process. The retrieval process is divided into three key phases: evidence merging, forward retrieval, backward update, and reasoning.

\subsubsection{Forward Retrieval}

Given that the agent holds a trustworthiness value $T^t(p_1)$ towards player $p_1$, if there exists a evidence chain $\mathcal{C}_n: p_o \rightarrow p_{o-1} \rightarrow \dots \rightarrow p_1$, the value $V_{\mathcal{C}_n}$ of this evidence chain and the cumulative trustworthiness update $u^t_i(p_o)$ towards player $p_o$ are computed as follows:

\begin{equation}\label{eqt:5}
	V_{\mathcal{C}_n} = \sum_{k=1}^{o-1} T^t(p_{k+1}) \cdot T^t(p_{k+1}, p_k)
\end{equation},

\begin{equation}\label{eqt:6}
	u^t(p_o) = T^t(p_1) \cdot \prod_{k=1}^{o-1} T^t(p_{k+1}, p_k)
\end{equation}.

The uncertainty associated with the chain $\mathcal{C}_n$ is defined by:

\begin{equation}\label{eqt:7}
	H(\mathcal{C}_n) = -u^t(p_o) \log_2 u^t(p_o)
\end{equation}.

For the player $p_o$ with $m$ related evidence chains $\mathcal{C}_1, \mathcal{C}_2, \dots, \mathcal{C}_m$, the updated trustworthiness $T^t_i(p_o)$ is given by:

\begin{equation}\label{eqt:8}
	T^{t+1}(p_o) = \frac{\sum_{n=1}^{m} (V_{\mathcal{C}_n} - H(\mathcal{C}_n)) \cdot u^t(p_o)}{\sum_{n=1}^{m} (V_{\mathcal{C}_n} - H(\mathcal{C}_n))}
\end{equation}
where the trustworthiness update is a weighted sum of the relevant evidence chains, where each chain’s weight is determined by its value and associated uncertainty.

\subsubsection{Backward Update}

Once $T^t(p_o)$ is updated, the edge weights associated with the relevant evidence chains need to be updated in reverse:

\begin{equation}\label{eqt:9}
	T^{t+1}(p_o, p_{o-1}) = \gamma \cdot \frac{T^t(p_o)}{T^t(p_{o-1})} + T^t(p_o, p_{o-1})
\end{equation}

Here, $\gamma$ represents the learning rate for the backward update, and $p_{o-1}$ is the preceding player in the evidence chain $\mathcal{C}_n$ $(n = 1,2,\dots m)$.


\subsubsection{Reasoning}

After updating the trustworthiness of the agent towards $p_o$, A summary and reasoning are made based on the trustworthiness of player $p_o$ and the relevant evidence chains retrieved.  Specifically, the trustworthiness of the agent towards $p_o$ and the evidence chains are combined into a prompt sent to LLM, which ultimately returns the summary and reasoning of the player $p_o$. The prompt used is shown in Appendix 1.2.

\begin{algorithm}[h]
	\caption{Graph Retrieval Augmented Reasoning}
	\label{alg:reasoning}
	\textbf{Input}: Number of the selected top trustworthiness nodes $w$, the target player $p_o$;\\
	\textbf{Initialization}: Players $p_1, \dots, p_n$; Nodes $N$ in $G^t_i$; Evidence chains list $C \leftarrow [\mathcal{C}_1, \mathcal{C}_2, \dots, \mathcal{C}_w]$ for player $p_o$ (initially empty); Priority Queue $Q \leftarrow \emptyset$;
	\begin{algorithmic}[1]
		\STATE $N \leftarrow \text{Sort}(N, T^t(n))$; // Sort $N$ by trustworthiness.
		\STATE $\{n_1, n_2, \dots, n_w\} \leftarrow \text{Top-}w(N)$; // Select the top $w$ nodes.
 		\STATE $Q \leftarrow \{n_1, n_2, \dots, n_w\}$;
 		\FOR{$j = 1, 2, \dots, w$}
			\STATE $\mathcal{C}_j \leftarrow \{n_j\}$;
		\ENDFOR
		\WHILE{$Q \neq \emptyset$}
			\STATE $n_c, \mathcal{C}_c \leftarrow \textbf{argmax}_{n \in Q} \, T^t_i(n)$;
			\STATE $Q \leftarrow Q \setminus \{n_c\}$;
			\FOR{each $n_k \in \text{Neighbors}(n_c)$}
				\STATE Merge evidence $e^t(p_k, p_c)$ to update $T^{t+1}(p_k, p_c)$ based on the Eq. \eqref{eqt:4}; // $p_k$, $p_c$ are the players corresponding to the nodes $n_k$, $n_c$.
			\ENDFOR
			\STATE $n_{k^*} \leftarrow \textbf{argmax}_{n_k \in \text{Neighbors}(n_c)} T^t(p_k)$;
			\STATE $\mathcal{C}_c \leftarrow \mathcal{C}_c \cup \{n_{k^*}\}$;
			\STATE $Q \leftarrow Q \cup \{n_{k^*}\}$;
 		\ENDWHILE
 		\STATE Use $C$ to update $T^{t+1}(p_o)$ based on the Eqs. \eqref{eqt:5}\eqref{eqt:6}\eqref{eqt:7}\eqref{eqt:8};
     	\STATE Update $T^{t+1}(p_o,p_{o-1})$ based on the Eq. \eqref{eqt:9};
 		\STATE Summarize and reason based on $T^t(p_o)$ and the evidence chains retrieved $C$;
	\end{algorithmic}
\end{algorithm}

\section{Experiments}
In this section, we evaluate the enhancement of LLMs' reasoning and intent analysis capabilities with GRATR, testing it on both the \emph{Werewolf} game and the Twitter dataset from the 2024 U.S. election. We use pure LLMs as the baseline, alongside state-of-the-art algorithms, including NativeRAG \cite{lewis2020retrieval}, RerankRAG \cite{sun2023chatgpt}, and LightRAG \cite{guo2024lightrag}, for comparison. 


\subsection{Experiment on Werewolf Game}

We implemented our GRATR method using the classic multiplayer game \emph{Werewolf} \cite{xu_exploring_2024}. The game consists of 8 players, including three leaders (the witch, the guard, and the seer), three werewolves, and two villagers. The history message window size $K$ is set to 15. We use the GPT-3.5-turbo, GPT-4o, GPT-4o-mini, Qwen-Max, and DeepSeek-V3 models as the backend LLMs, with their temperatures set to 0.3. In each game, four players are assigned to each algorithm, with three players randomly assigned to the leader and werewolf roles, and the remaining player assigned to the village side. The algorithm corresponding to the winning side is considered the winner of the game. Each algorithm participated in 50 games with different backend LLMs.

\subsubsection{Win Rate Analysis}

Table \ref{tab:wr} presents the win rates of LLMs with GRATR in pairwise comparisons against the baseline and LLMs with NativeRAG, RerankRAG, and LightRAG in the \emph{Werewolf} game. The win rates include total win rate (TWR), werewolf win rate (WWR), and leader win rate (LWR).

\begin{table}[h!]
\centering
\resizebox{\linewidth}{!}{
\begin{tabular}{lcccc}
\toprule
GRATR vs.  & Model & TWR & WWR & LWR \\
\midrule
\multirow{5}{*}{\makecell{Baseline}}
    & GPT-3.5-turbo & 76.0\% & 72.0\% & 80.0\% \\
    & GPT-4o & 88.0\% & 84.0\% & 92.0\% \\
    & GPT-4o-mini & 84.0\% & 76.0\% & 92.0\% \\
    & Qwen-max & 94.0\% & 88.0\% & 100.0\% \\
    & DeepSeek-v3 & 92.0\% & 88.0\% & 96.0\% \\
    \cmidrule{2-5}
    & Mean & 86.8\% & 81.6\% & 92.0\% \\
\midrule
\multirow{5}{*}{\makecell{NativeRAG}}
    & GPT-3.5-turbo & 66.0\% & 60.0\% & 72.0\% \\
    & GPT-4o & 80.0\% & 76.0\% & 84.0\% \\
    & GPT-4o-mini & 78.0\% & 72.0\% & 84.0\% \\
    & Qwen-max & 80.0\% & 76.0\% & 84.0\% \\
    & SeepAeek-v3 & 88.0\% & 80.0\% & 96.0\% \\
    \cmidrule{2-5}
    & Mean & 78.4\% & 72.8\% & 84.0\% \\
\midrule
\multirow{5}{*}{\makecell{RerankRAG}}
    & GPT-3.5-turbo & 72.0\% & 76.0\% & 68.0\% \\
    & GPT-4o & 90.0\% & 84.0\% & 96.0\% \\
    & GPT-4o-mini & 80.0\% & 80.0\% & 80.0\% \\
    & Qwen-max & 92.0\% & 84.0\% & 100.0\% \\
    & DeepSeek-v3 & 90.0\% & 84.0\% & 96.0\% \\
    \cmidrule{2-5}
    & Mean & 84.8\% & 81.6\% & 88.0\% \\
\midrule
\multirow{5}{*}{\makecell{LightRAG}}
    & GPT-3.5-turbo & 80.0\% & 76.0\% & 84.0\% \\
    & GPT-4o & 90.0\% & 96.0\% & 84.0\% \\
    & GPT-4o-mini & 84.0\% & 80.0\% & 88.0\% \\
    & Qwen-max & 88.0\% & 84.0\% & 90.0\% \\
    & DeepSeek-v3 & 88.0\% & 96.0\% & 80.0\% \\
    \cmidrule{2-5}
    & Mean & 86.0\% & 86.4\% & 85.2\% \\
\bottomrule
\end{tabular}
}
\caption{The total, werewolf, and leader win rates (TWR, WWR, LWR) of LLMs with GRATR in pairwise comparisons against the baseline and LLMs with NativeRAG, RerankRAG, and LightRAG in the \emph{Werewolf} game.}
\label{tab:wr}
\end{table}

From the mean TWR in Table \ref{tab:wr}, it is clear that GRATR significantly outperforms both pure LLMs and LLMs with advanced RAG methods. Except for the match against NativeRAG, where the win rate is 78.4\%, GRATR achieves win rates above 80\% in all other pairwise competitions. The experimental results support the claim that GRATR effectively enhances LLM reasoning in incomplete information games and improves win rates. More specifically, the results show that GRATR achieves the highest win rate when playing against pure LLMs, followed by LightRAG, RerankRAG, and NativeRAG. This suggests that external retrieval-based techniques are beneficial for enhancing LLM reasoning. Furthermore, while LightRAG, as a graph-based retrieval-augmented generation method, excels at summarization rather than reasoning, and RerankRAG, though a commendable variant, fails to capture the causal relationships of player actions in multi-hop retrieval, which results in its lower performance compared to NativeRAG.

Further analysis of the win rates when the agent plays as a werewolf or leader reveals that the agent performs significantly better as a leader. Notably, when using Qwen-Max, the win rate reaches 100\% against both Baseline and RerankRAG. This can be attributed to the game dynamics of \emph{Werewolf}, where the werewolf must deceive other players to conceal their identity, whereas the leader only needs to reason out who the werewolf is. The high win rates for the leader role provide evidence that GRATR enhances the reasoning ability of LLMs, enabling them to identify the concealed werewolf. Although GRATR also performs well when the agent plays as a werewolf, the deception required for this role presents a greater challenge for LLMs.

The experiment shows that different LLMs have a significant impact on the results. As shown in Table \ref{tab:wr}, GRATR achieves better performance with GPT-4o, Qwen-Max, and DeepSeek-V3, with win rate improvements of 4\%, 2\%, 10\%, and 4\%, respectively. Publicly available evidence \cite{chiang2024chatbot,2023opencompass} indicates that GPT-4o, Qwen-Max, and DeepSeek-V3 exhibit stronger reasoning capabilities compared to GPT-3.5-turbo and GPT-4o-mini. Therefore, we conclude that stronger LLMs further amplify the performance advantages of GRATR.

\subsubsection{Action Scores}

In the \emph{Werewolf} game, win rate alone evaluates the overall performance of the team, but it does not fully reflect the individual agent's actual performance. Therefore, this section further analyzes the agent's action scores in each game to highlight its superiority in reasoning, social interaction, role identification, and other aspects. Table \ref{tab:performance-calc} presents the detailed scoring breakdown for agents under different identities, including scores for correct and incorrect votes. Additionally, each winning player is pre-allocated a base score of 5 points.

\begin{table}[htbp]
\centering
    \resizebox{\linewidth}{!}{
	\begin{tabular}{cccccc}
		\toprule
	   & \textbf{Werewolf} & \textbf{Witch} & \textbf{Guard} & \textbf{Seer} & \textbf{Villager} \\
		\midrule
		\textbf{Correct} & 0.5   & 1.5   & 1.5   & 1.5   & 1.0  \\
		\textbf{Incorrect} & -0.5  & -1.5  & -1.5  & -1.5  & -1.0 \\
		\bottomrule
	\end{tabular}%
    }
    	\caption{Correct and incorrect action scores for different identities in Werewolf game}

	\label{tab:performance-calc}%
\end{table}

\begin{figure*}[htbp]
	\centering
	\begin{minipage}[t]{0.25\linewidth}
		\centering
		\includegraphics[width=\textwidth]{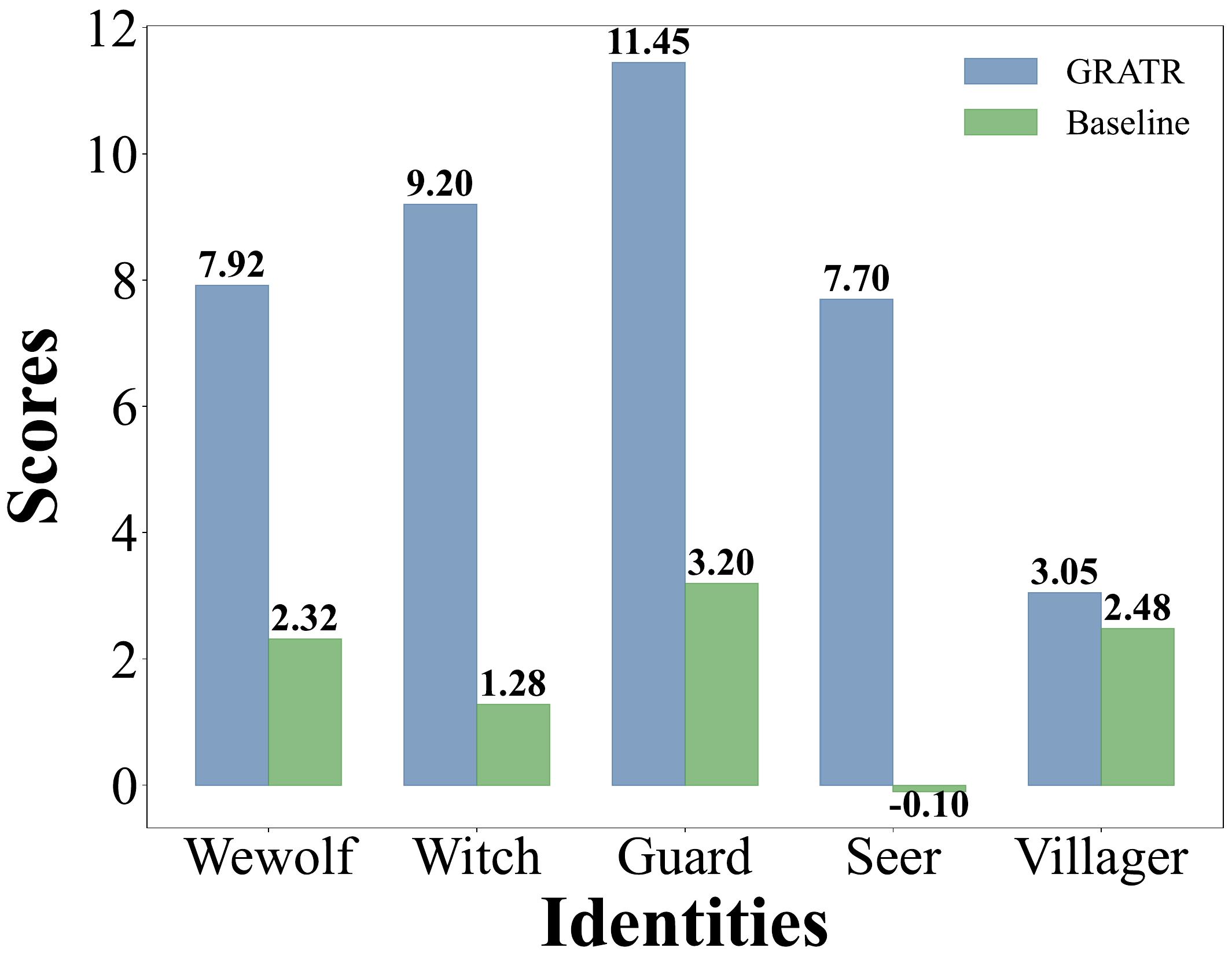}
		\subcaption{GRATR vs. baseline.}
		\label{fig:performance1}
	\end{minipage}%
	\hfill
	\begin{minipage}[t]{0.25\linewidth}
		\centering
		\includegraphics[width=\textwidth]{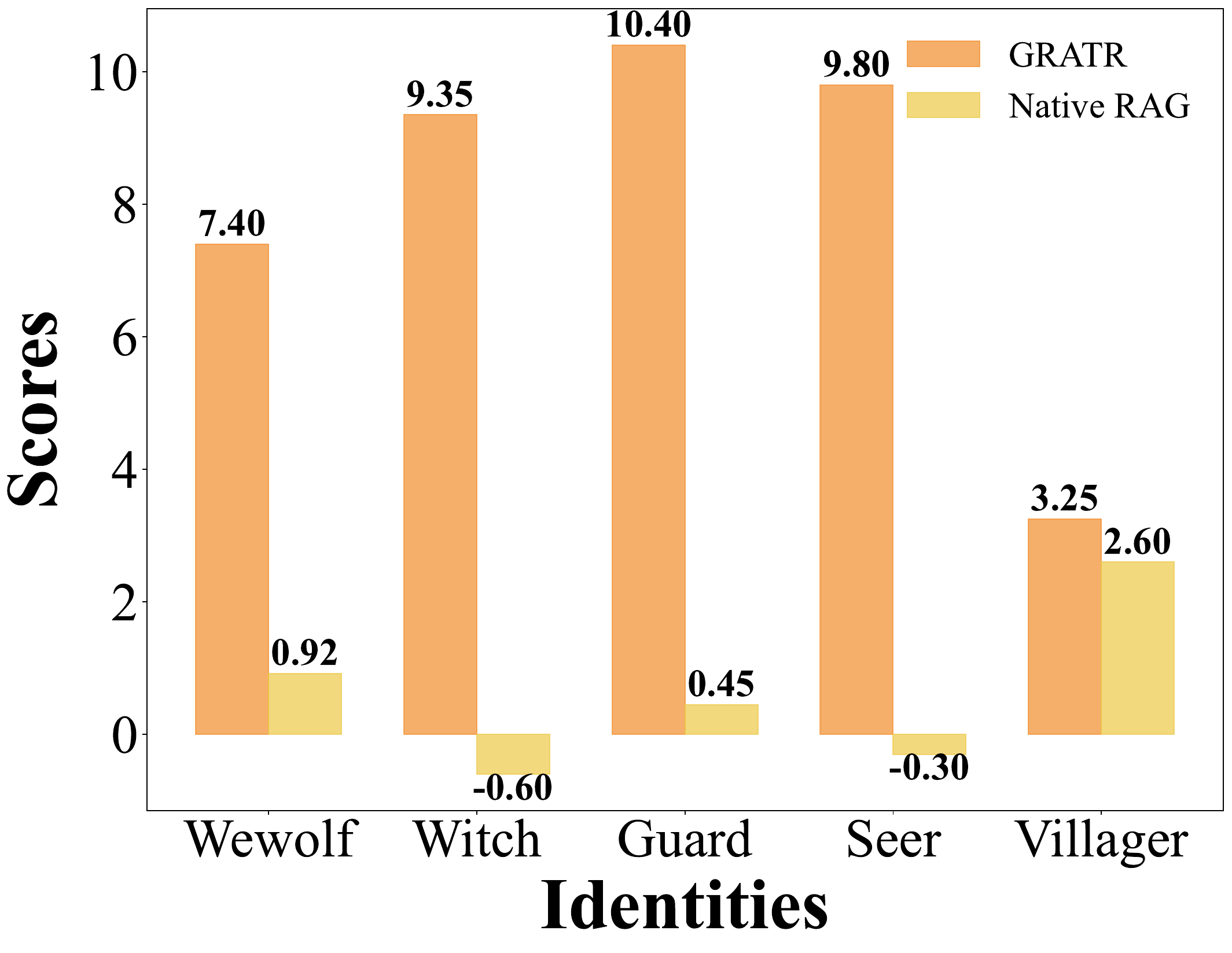}
		\subcaption{GRATR vs. NativeRAG.}
		\label{fig:performance2}
	\end{minipage}%
	\hfill
	\begin{minipage}[t]{0.25\linewidth}
		\centering
		\includegraphics[width=\textwidth]{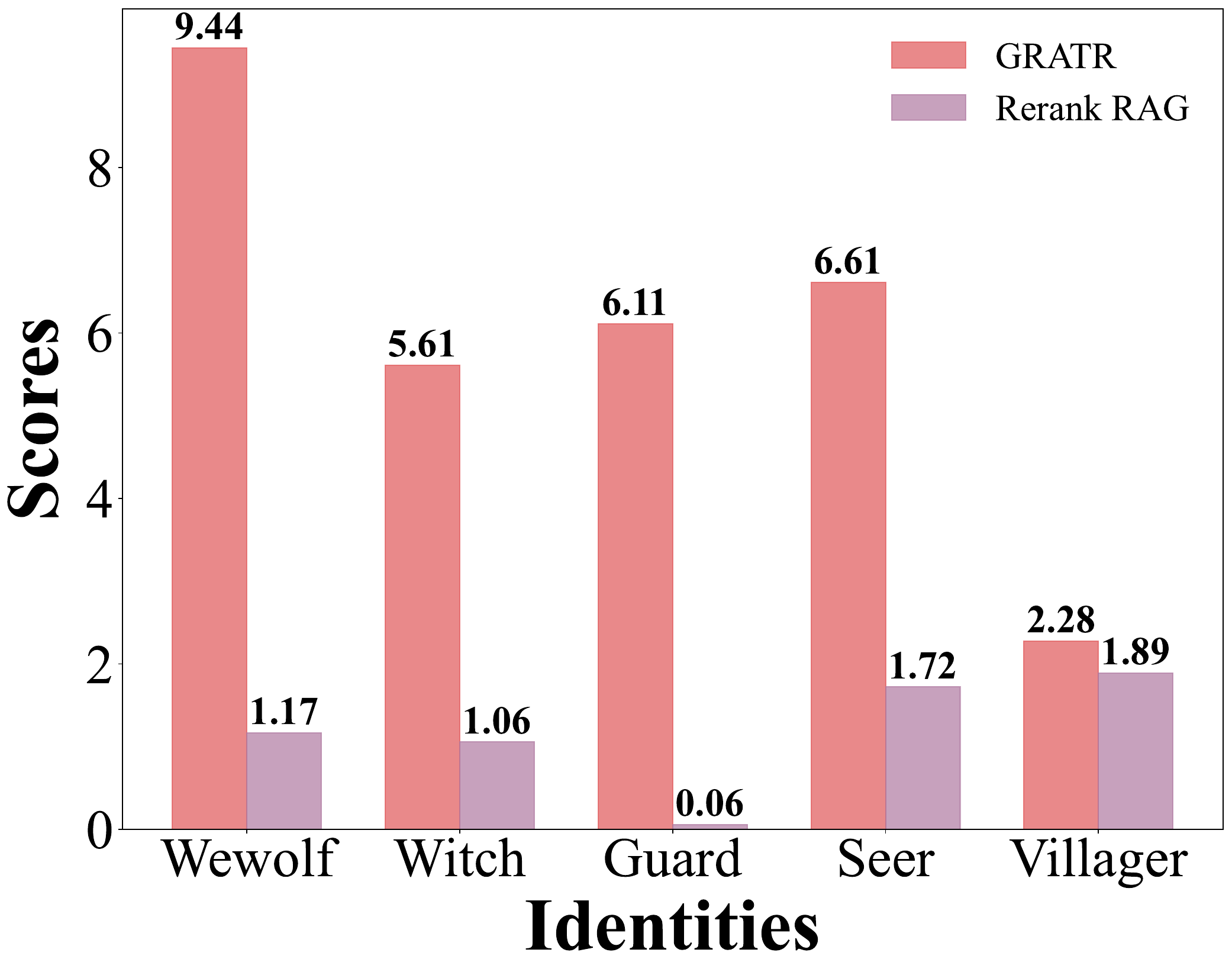}
		\subcaption{GRATR vs. RerankRAG.}
		\label{fig:performance3}
	\end{minipage}%
	\hfill
	\begin{minipage}[t]{0.25\linewidth}
		\centering
		\includegraphics[width=\textwidth]{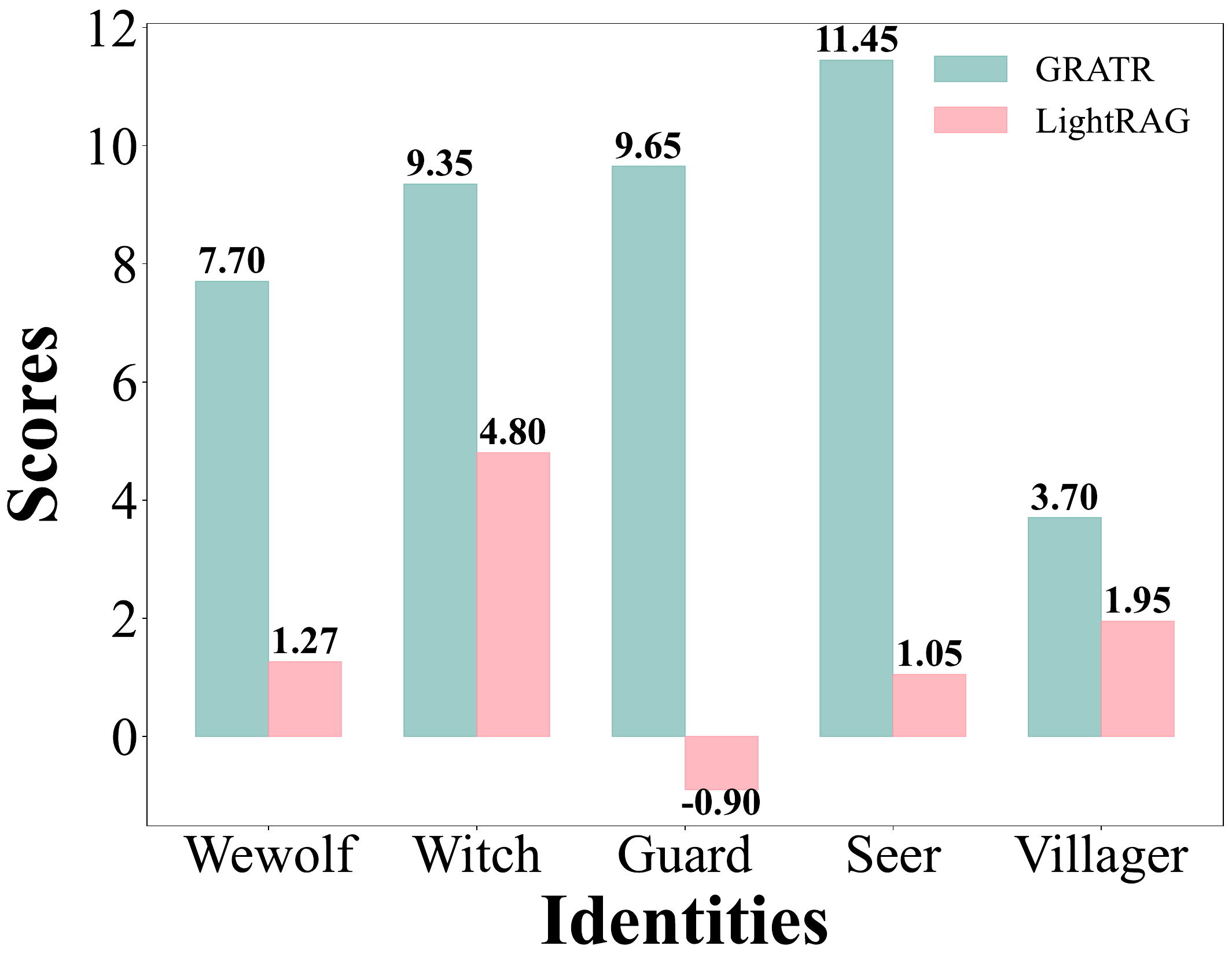}
		\subcaption{GRATR vs. LightRAG.}
		\label{fig:performance4}
	\end{minipage}
	\caption{Action scores of GRATR vs. baseline LLM, LLM with NativeRAG, RerankRAG, and LightRAG in Werewolf game}
	\label{fig:combined_performance}
\end{figure*}

Fig. \ref{fig:combined_performance} presents the action scores of GRATR vs. baseline LLM, LLM with NativeRAG, RerankRAG, and LightRAG in the \emph{Werewolf} game. Overall, when the agent plays as a villager, the score differences are minimal, generally under 2 points, and even less than 1 point when compared to baseline LLM and LLMs with NativeRAG or RerankRAG. This indicates that, on average, the agent makes fewer than one error per game round. It is important to note that in the \emph{Werewolf} game, villagers have no prior information other than their own identity, so all reasoning is based on the inconsistencies and consistencies in players' actions rather than validating with prior knowledge. Therefore, the superior behavior scores of GRATR when the agent plays as a villager demonstrate the algorithm's multi-hop retrieval capability and its advantage in causal reasoning.

For identities other than the villager, such as the werewolf and leader, the action score differences are significantly larger. A major portion of this difference stems from the win rate, as the winning side is awarded a base score of 5 points. The remaining differences are due to the correctness of the agent's actions. For example, when the agent plays as a werewolf, the score difference is greater than 5 points, indicating that the agent made incorrect actions. However, in the \emph{Werewolf} game, the werewolf has prior knowledge of all teammates and opponents, so any errors in actions are primarily attributed to LLM hallucination. While there may be potential deception and disguise involved, this section does not delve further into this aspect, as no additional supporting information is available.

\subsubsection{Hallucination Detection}

This section detects LLM hallucination by analyzing the agent's thinking and actions. If the agent's reasoning aligns with their true stance, the process is correct. If the agent's actions conflict with their reasoning, it indicates strategic deception. If the agent's thinking deviates from their real stance, it signals cognitive bias or hallucination. We manually labeled the consistency of the agent's thinking and actions, with the results shown in Table \ref{tab:pr-hall}.

\begin{table}[htbp]
\centering
\resizebox{\linewidth}{!}{
\begin{tabular}{lcccc}
\toprule
\textbf{Method} & \textbf{Identity} & \textbf{Correct Reasoning} & \textbf{Deception} & \textbf{Hallucination} \\
\midrule
Baseline        & Werewolf          & 61.9\%                     & 1.5\%            & 36.6\%                 \\
                & Leader            & 69.7\%                     & 0.5\%            & 29.8\%                 \\
\midrule
GRATR           & Werewolf          & \textbf{85.1}\%                     & \textbf{11.4}\%           & \textbf{3.5}\%                  \\
                & Leader            & \textbf{97.0}\%                     & 1.3\%            & \textbf{1.7}\%                  \\
\midrule
NativeRAG      & Werewolf          & 79.1\%                     & 3.1\%            & 17.8\%                 \\
                & Leader            & 84.4\%                     & 1.7\%            & 13.9\%                 \\
\midrule
RerankRAG      & Werewolf          & 74.6\%                     & 9.4\%            & 16.0\%                 \\
                & Leader            & 83.9\%                     & \textbf{6.0}\%            & 10.1\%                 \\
\midrule
LightRAG        & Werewolf          & 76.2\%                     & 10.1\%           & 13.7\%                 \\
                & Leader            & 86.2\%                     & 4.5\%            & 9.3\%                  \\
\bottomrule
\end{tabular}
}
\caption{Correct Reasoning, deceive, and hallucination rates of different methods in Werewolf and Leader identities.}
\label{tab:pr-hall}
\end{table}

The data in the table shows that the GRATR method outperforms others in both correct reasoning and hallucination mitigation. It improves correct reasoning by at least 6\% and 12.6\% for the Werewolf and Leader identities, respectively. It also exhibits an 11.4\% deception rate for the Werewolf identity. However, the deception rate is lower for the Leader identity, as the Leader typically needs to reveal their identity to guide the villagers to victory, with deception used only for strategic purposes. Most importantly, GRATR significantly mitigates LLM hallucination, reducing them by a factor of 10 for the Werewolf identity and 17 for the Leader identity compared to the baseline. These results strongly support GRATR’s effectiveness in enhancing LLM reasoning capabilities and reducing hallucination.

\subsection{Experiment on Intent Analysis}
In this section, we utilize a public dataset of Twitter tweets at the time of the U.S. election \cite{balasubramanian2024public} to evaluate GRATR for their intent analysis capability. This dataset comprehensively captures large-scale social media discourse related to the 2024 U.S. presidential election. The dataset includes approximately 27 million publicly available political tweets collected between May 1 and November 1, 2024. Each tweet is accompanied by detailed metadata, including precise timestamps and multi-dimensional user engagement metrics (such as reply count, retweet count, like count, and view count).

\subsubsection{Experimential Results}
We define five possible tweet intents: Anti-Democrat, Anti-Republican, Pro-Democrat, Pro-Republican, and Neutral \cite{ibrahim2024analyzing}. To evaluate the accuracy of the algorithm, we manually annotated the intents of 26,523 valid tweets (i.e., tweets that are not garbled and are meaningful). For this, we generalized each tweet to the individual who sent it (since a person may have sent multiple tweets) and followed the timeline to simulate a real Twitter discussion environment. In this setup, we applied LLMs with GRATR, baseline LLMs, LLMs with NativeRAG, RerankRAG, and LightRAG to analyze the stance of each individual and further analyze the intent of their tweets. Table \ref{tab:ele} presents the accuracy and macro F1-score of all comparison algorithms for intent analysis of the tweets.

\begin{table}[htbp]
\centering
\resizebox{\linewidth}{!}{
\begin{tabular}{lccccc}
\toprule
 & \textbf{Baseline} & \textbf{GRATR} & \textbf{NativeRAG} & \textbf{RerankRAG} & \textbf{LightRAG} \\
\midrule
\textbf{Accuracy} & 0.818 & \textbf{0.922} & 0.868 & 0.879 & 0.891 \\
\textbf{Macro F1} & 0.809 & \textbf{0.914} & 0.869 & 0.878 & 0.893 \\
\bottomrule
\end{tabular}
}
\caption{Accuracy and Macro F1-score of baseline LLMs, LLMs with GRATR, NativeRAG, RerankRAG, and LightRAG on intent analysis of the tweets.}
\label{tab:ele}
\end{table}

Among all the methods, LLMs enhanced with GRATR achieve the highest accuracy of 0.923 and a macro F1-score of 0.918, demonstrating superior performance. The accuracy metric reflects the proportion of correctly classified tweets out of the total. However, accuracy alone may not fully capture performance when dealing with imbalanced data, such as political tweets, where certain intents (e.g., Pro-Democrat or Anti-Republican) are more prevalent than others. In these cases, the macro F1-score provides a more balanced evaluation by considering both precision and recall for each intent category individually, ensuring equal weight for less frequent categories. The significantly higher macro F1-score of LLMs with GRATR (0.918), compared to the baseline models (0.762), indicates that GRATR enhances the model’s ability to accurately predict all intents, especially the subtle or less frequent ones, in politically charged discourse. This result highlights GRATR’s capacity to integrate contextual and temporal information, which is critical for understanding the nuanced intents of tweets, particularly in dynamic environments like social media during a presidential election. Additionally, LLMs with RAG, including NativeRAG, RerankRAG, and LightRAG, all outperform the baseline LLMs, underscoring the effectiveness of RAG in improving intent analysis.

\section{Related Work}
\textbf{Reasoning Task.} In incomplete information games, players enhance decision-making by reasoning through observed data and analyzing behaviors in real time, despite misleading information \cite{wu_enhance_2024,zhang_controlling_2024,cheng_self-playing_2024,qin_arena_2024,costarelli_gamebench_2024}. Traditional methods like Bayesian approaches \cite{zamir2020bayesian}, evolutionary game theory \cite{deng2015evidence}, and machine learning techniques such as Monte Carlo tree search \cite{cowling2012information} and reinforcement learning (RL) \cite{heinrich2016deep} have been used, with RL gaining prominence for its inference capabilities. However, RL's reliance on domain-specific data limits generalizability. Large language models (LLMs) offer an alternative with extensive knowledge and language capabilities, as shown by Xu et al. \cite{xu2023language}, who combined LLMs and RL for strategic language agents. Yet, LLMs face challenges like high training costs, inability to update data in real time, and hallucination, hindering real-time reasoning in multiplayer games. RAG addresses these limitations, enhancing LLMs' reasoning in dynamic game environments.

\textbf{Retrieval Augmented Generation.} RAG enhances LLMs by integrating external knowledge retrieval. NativeRAG \cite{lewis2020retrieval} involves document chunking/encoding, vector-based semantic retrieval, and prompt construction. While efficient, it often retrieves low-relevance chunks. RerankRAG improves accuracy by adding a reranking step (e.g., transformer-based cross-encoders) to prioritize relevant chunks \cite{sun2023chatgpt}. GraphRAG uses knowledge graphs, modeling entities as nodes and relationships as edges, supporting multi-hop reasoning, and capturing complex dependencies for deeper queries \cite{edge2024local}. Both Rerank and GraphRAG increase computational complexity. LightRAG \cite{guo2024lightrag} mitigates this with lightweight strategies like heuristic filtering, balancing efficiency and relevance. Retrieval-Augmented Reasoning (RAR) \cite{tran2024rare} integrates dynamic knowledge retrieval with reasoning modules, improving temporal relevance but facing challenges in multi-step inference and trustworthiness verification.

\section{Conclusion}
This paper has presents a novel RAG framework, named GRATR, which utilizes a dynamic trustworthiness graph that is updated in real-time with new evidence, to enhance the accuracy of trustworthiness assessments. The framework consists of two main phases. During the agent observation phase, evidence is collected to update the nodes and edges of the graph. During the agent's turn, relevant evidence chains are retrieved to assess the trustworthiness of the player's actions, thereby improving reasoning and decision-making. Experiments conducted in the multiplayer game \emph{Werewolf} demonstrate that GRATR outperforms existing methods in terms of game winning rate, overall performance, and reasoning ability, while mitigating LLM hallucination. Additionally, GRATR enables the traceability and visualization of the reasoning process through time-based evidence and evidence chains. Furthermore, GRATR’s application to the U.S. election Twitter dataset highlights its effectiveness in intent analysis, showcasing its potential for real-world applications.

\bibliographystyle{named}
\bibliography{ijcai25}

\end{document}